%% file: arxiv.tex
\documentclass[letterpaper]{article}
\usepackage[draft]{aaai2026}
\input{shared/preamble}

\nocopyright

\title{GRAFT: Gain-Recalibrated Adapters for Transformer-Based Neural Population Activity Modeling}

\author{
    Xiangsheng Ge\textsuperscript{\rm 1},
    Yang Xie\textsuperscript{\rm 1}\thanks{Corresponding author.}
}

\affiliations{
    \textsuperscript{\rm 1}Shanghai Jiao Tong University\\
    gxstjyh@gmail.com, yang.xie@sjtu.edu.cn
}

\begin{document}

\maketitle

\begin{abstract}
\input{sections/abstract}
\end{abstract}

\begin{links}
    \link{Code}{https://github.com/GTJYH/GRAFT}
\end{links}

\input{sections/main}


\bibliography{graft}

\clearpage
\onecolumn
\appendix
\input{sections/appendix}

\end{document}

%% file: shared/preamble.tex
\usepackage{times}  
\usepackage{helvet}  
\usepackage{courier}  
\usepackage[hyphens]{url}  
\usepackage{graphicx} 
\usepackage{amsmath}
\usepackage{amssymb}
\usepackage{natbib}
\usepackage{caption}
\usepackage{booktabs}
\usepackage{multirow}
\input{shared/macros}

%% file: shared/macros.tex
\newcommand{\method}{GRAFT}
\newcommand{\dataset}{NLB'21 MC Maze}
\newcommand{\cobps}{co-bps}
\newcommand{\fpbps}{fp-bps}
\newcommand{\velrtwo}{vel $R^2$}
\newcommand{\psthrtwo}{PSTH $R^2$}

%% file: sections/abstract.tex
Neural population activity models can recover rich temporal structure from binned spikes, but their read-in and readout layers often remain tied to a fixed set of recorded neurons. This coupling limits reuse in long-term brain-computer interfaces, where recorded neuron identities, counts, and response statistics can change across days. We introduce \method{}, a Transformer-based neural population activity model that separates reusable temporal dynamics from a recalibratable neuron interface. The neuron interface controls how recorded neurons enter and leave the shared backbone, and auxiliary gain and positional mechanisms support neural activity modeling inside the Transformer. On MC Maze under the standard NLB'21 protocol, \method{} reaches $0.3866$ \cobps{} as an ensemble, setting a new state of the art on the primary \cobps{} metric among public and reported NLB'21 results. In a cross-day protocol constructed from the \dataset{} dataset series, \method{} recalibrates from MC Maze to the scaled MC Maze datasets (Large/Medium/Small) by updating only $9.21\%$ of parameters, reaching $0.3749$, $0.3112$, and $0.3152$ \cobps{} with restricted target-day support sets. These results show that the same interface-backbone separation supports both strong Transformer-based neural population activity modeling and data-efficient cross-day recalibration.

%% file: sections/main.tex
\input{sections/introduction}
\input{sections/related_work}
\input{sections/problem_setup}
\input{sections/method}
\input{sections/experiments}
\input{sections/results}
\input{sections/discussion}
\input{sections/conclusion}

%% file: sections/introduction.tex
\section{Introduction}

Neural population activity modeling aims to recover firing-rate structure and temporal dynamics from noisy binned spikes. It is a central tool for studying neural computation and for building brain-computer interfaces (BCIs), where models must extract behaviorally relevant structure from partially observed neural populations \cite{Saxena2019PopulationDoctrine,Vyas2020Computation}. Recent latent-variable and Transformer-based models have become strong within-session firing-rate estimation and masked neural reconstruction baselines on standardized benchmarks such as NLB'21 \cite{Keshtkaran2022AutoLFADS,Pei2021NLB,Ye2021NDT,Le2022STNDT}. Strong within-session performance, however, does not by itself make a model reusable across recording days.

Long-term BCI use breaks a common assumption behind many neural population models: that the input and output coordinates refer to a fixed set of recorded neurons. In chronic recordings, electrode-tissue relationships, signal quality, recorded neuron identities, neuron counts, and firing statistics can change over time \cite{Orsborn2014CLDA,Fan2023PlugAndPlayStability,Pun2024Instability}. The difficulty goes beyond shifted target-day spike-count statistics. The model interface itself often assumes that the same coordinate refers to the same recorded neuron, an assumption that long-term recordings do not reliably satisfy. If the read-in and readout layers are tied to source-day neuron indices, reusing a source-day model on a target day with a different recorded population becomes difficult. Fully retraining the model for each target day is also poorly matched to BCI calibration, where target-day data and calibration time are limited.

We introduce \method{}, a Transformer-based neural population activity model that treats temporal dynamics and the neuron interface as different modeling objects. \method{} uses a reusable Transformer dynamics backbone to model temporal structure after neural activity has been mapped into a fixed-dimensional representation. A neuron interface controls how recorded neurons enter and leave this shared backbone. Each recorded neuron is represented by a learnable embedding; the embedding parameterizes both the read-in gain used to inject spike counts into the backbone and the readout direction used to predict that neuron's firing-rate parameter. Changes in neuron count or identity are therefore handled at the interface level, while the backbone dimension remains fixed.

Several lightweight mechanisms make this interface-backbone separation effective in the evaluated setting. Value-side attention gain modulates the effective strength of aggregated context without changing query-key matching inside the Transformer. Neural positional encoding combines trial-stage position with local temporal-distance information, which matches the structured preparation and movement phases of MC Maze trials. Interface-level contrastive consistency stabilizes read-in representations under masking perturbations, and repeated masking exposes the same few support trials under multiple mask patterns during target-day recalibration.

To evaluate this design, we construct a restricted-support cross-day recalibration protocol from the \dataset{} dataset series. MC Maze is used as the source day, and the scaled MC Maze datasets (Small, Medium, and Large) are used as target days. NLB'21 introduced these datasets for data-scaling evaluation, but each comes from a separate experimental session with a different recording date and neuron count, making them a natural testbed for cross-day recalibration within a matched task family. The protocol preserves task comparability while introducing recording-day changes in the observed population. Target-day support trials are restricted; validation data are used for checkpoint and ensemble selection; the public test split is reserved for final evaluation.

On MC Maze under the standard NLB'21 protocol, the \method{} ensemble reaches $0.3866$ \cobps{}, setting a new state of the art on the primary \cobps{} metric among public and reported results. In cross-day recalibration, \method{} reuses the source-day Transformer backbone and updates only the neuron interface, corresponding to about $9.2\%$ of model parameters, while using about $61\%$ of the target-day training split on average. Under this restricted setting, it exceeds the reported target-day \cobps{} of strong full-data baselines such as AutoLFADS, supporting the role of the neuron interface in data- and parameter-efficient recalibration.

Our contributions are:
\begin{itemize}
    \item We introduce \method{}, a Transformer-based neural population activity model that separates reusable temporal dynamics from a recalibratable neuron interface, combining neuron-conditioned read-in/readout with value-side attention gain, neural positional encoding, and interface-level contrastive consistency; on MC Maze under the standard NLB'21 protocol, \method{} achieves state-of-the-art \cobps{}.
    \item We construct a restricted-support cross-day recalibration protocol from the \dataset{} dataset series, using MC Maze as the source day and the scaled MC Maze datasets (Small/Medium/Large) as naturally different target days, with target-day support sets averaging about $61\%$ of the target-day training splits.
    \item We develop an interface-only cross-day recalibration procedure that freezes the source-day Transformer backbone and updates target-day neuron-interface parameters using repeated masking, adapting about $9.2\%$ of model parameters while exceeding several baselines trained with full target-day data.
\end{itemize}

%% file: sections/related_work.tex
\section{Related Work}

\paragraph{Neural Population Activity Modeling.} Latent-variable models have long been used to infer smoother firing-rate structure and lower-dimensional dynamics from noisy neural population activity. LFADS formulates single-trial neural activity as a sequential latent dynamical system and maps latent trajectories back to neuron-wise firing rates \cite{Pandarinath2018LFADS}. AutoLFADS addresses the training sensitivity of this family by automating regularization schedules and hyperparameter search, making it a competitive baseline for firing-rate estimation across neural datasets \cite{Keshtkaran2022AutoLFADS}. NLB'21 further standardized this evaluation by specifying train/validation/test splits, held-out neurons, future time windows, and metrics such as \cobps{}, \fpbps{}, \psthrtwo{}, and \velrtwo{} \cite{Pei2021NLB}. These models and benchmarks provide the firing-rate modeling setting used here, but their common implementations are usually built around a fixed recorded population. The read-in and readout maps therefore remain coupled to the neuron set used during training.

\paragraph{Transformer Models for Neural Activity.} Transformer-based models avoid imposing an explicit recurrent latent state transition and instead use masked sequence modeling over binned neural activity. NDT adapts BERT-style masked reconstruction to neural population activity by using a Transformer encoder to estimate firing-rate parameters at masked positions \cite{Ye2021NDT}. STNDT extends this idea with spatiotemporal structure, allowing the model to capture temporal context and interactions among neurons within the same recording session \cite{Le2022STNDT}. These models motivate our use of a Transformer backbone for neural population activity modeling. Their interfaces, however, still operate on the neuron coordinates available in a given recording. When the recorded population changes across days, the temporal backbone and the neuron-indexed read-in/readout layers are not cleanly separated. \method{} keeps the Transformer as a reusable temporal modeling component, but changes the interface through which neurons enter and leave the backbone.

\paragraph{BCI Stability and Cross-Day Recalibration.} Long-term intracortical BCI systems must handle changes in recorded neural signals over time. Closed-loop decoder adaptation, plug-and-play stability, and chronic recording analyses show that neural interfaces can require recalibration as signal quality, recorded units, and neural response statistics change \cite{Orsborn2014CLDA,Fan2023PlugAndPlayStability,Pun2024Instability}. Recent work has addressed this instability through alignment, identity-aware modeling, or robust decoding benchmarks. NoMAD aligns target-day neural activity to a source-day latent dynamics model for stable decoding \cite{Karpowicz2025NoMAD}. SPINT treats neuron order and identity as variables rather than fixed coordinates, using context-dependent embeddings for consistent cross-session decoding \cite{Le2025SPINT}. FALCON evaluates few-shot neural decoding for robust iBCI settings \cite{Karpowicz2024FALCON}. These efforts are closely related in motivation, but their primary emphasis is neural decoding or cross-session alignment. Our evaluation follows an NLB'21-style neural population activity modeling setting: the question is whether a source-day activity model can reuse temporal structure across MC Maze recording days while recalibrating only the neuron-specific interface.

\paragraph{Gain Modulation and Multiplicative Interfaces.} Gain modulation offers a useful modeling principle for separating neuron identity from shared dynamics. In neuroscience, gain modulation describes changes in neural response amplitude as a function of task context or internal state \cite{Salinas2000Gain}, and context-dependent population dynamics provide one example of how the same neural population can support different computations under different conditions \cite{Mante2013ContextDependent}. We use this idea computationally rather than as a claim of biological mechanism: in \method{}, gain controls how neuron-specific spike counts enter a shared temporal backbone and how backbone states are read out for each neuron. Multiplicative control also appears in neural-network gates and gated attention mechanisms, where feature or attention outputs are rescaled before further computation \cite{Qiu2025GatedAttention}. \method{} applies this principle at two levels: the neuron interface modulates read-in and readout for recorded neurons, and attention gain modulates value-side context inside the Transformer backbone.

%% file: sections/problem_setup.tex
\section{Problem Setup}

\paragraph{NLB'21-Style Neural Population Activity Modeling.} We consider trial-wise neural population activity represented as binned spike counts. For a trial with $T$ time bins and $N$ recorded neurons, let
\begin{equation}
    \mathbf{Y} = [y_{t,n}] \in \mathbb{N}^{T\times N},
    \label{eq:problem-input}
\end{equation}
where $y_{t,n}$ is the spike count of neuron $n$ in time bin $t$. Following the NLB'21 formulation \cite{Pei2021NLB}, the model receives the visible part of a trial and predicts firing-rate parameters $\hat{\lambda}_{t,n}$ for specified held-out neurons or future time bins. We use a Poisson observation model for spike counts, so training losses and evaluation likelihoods are defined on firing-rate parameters rather than on deterministic reconstruction of observed counts.

The primary metric is \cobps{}, which measures how much the predicted firing-rate parameters improve the held-out-neuron likelihood over a neuron-wise null rate, normalized by spike count. For an evaluation set $\Omega_c$, it can be written as
\begin{equation}
    \begin{aligned}
    \mathrm{co\mbox{-}bps}
    &=
    \frac{\sum_{(t,n)\in\Omega_c}\Delta_{t,n}}
    {\log 2 \sum_{(t,n)\in\Omega_c} y_{t,n}},\\
    \Delta_{t,n}
    &=
    \log p(y_{t,n}\mid \hat{\lambda}_{t,n})\\
    &\quad-
    \log p(y_{t,n}\mid \lambda^{\mathrm{null}}_n).
    \end{aligned}
    \label{eq:co-bps}
\end{equation}
We report \cobps{} as the main metric because it directly evaluates whether observed neurons support recovery of held-out neural activity. We also report \fpbps{}, \psthrtwo{}, and \velrtwo{}. These metrics capture future-bin firing-rate prediction, condition-averaged firing structure, and behaviorally decodable information in the submitted rates, respectively. They are useful secondary views, but they do not always rank methods in the same order because they evaluate different aspects of the output.

\paragraph{MC Maze Modeling and Target-Day Recalibration.} The MC Maze setting evaluates whether \method{} is a strong neural population activity model under the standard NLB'21 protocol. The model is trained on the MC Maze training split, selected using validation data, and evaluated on the public test split with the standard NLB'21 metrics. This result measures within-day firing-rate modeling quality under the same evaluation target used by prior NLB'21 methods, while also providing the initialization for target-day recalibration.

The target-day setting evaluates whether a source-day model can be recalibrated when the recorded population changes. We use MC Maze recorded on 2009-09-25 as the source day, and the scaled MC Maze datasets (Small, Medium, and Large) as target days. These target datasets come from the same monkey and task family but have different recording dates and neuron counts: Small was recorded on 2009-09-28 with 107 visible and 35 held-out neurons, Medium on 2009-09-29 with 114 visible and 38 held-out neurons, and Large on 2009-10-06 with 122 visible and 40 held-out neurons. The observed and future time windows remain 140 and 40 bins. This construction keeps the behavioral task comparable while changing the neural interface the model must use.

Target-day recalibration is deliberately restricted. From the target-day training split, we use $48/75$ trials for Small, $96/188$ trials for Medium, and $256/375$ trials for Large, corresponding to support ratios of $64.0\%$, $51.1\%$, and $68.3\%$. The Transformer dynamics backbone learned on MC Maze remains frozen. Only the neuron interface is updated on target-day support trials. Target-day validation data are used for checkpoint and ensemble selection, and the public test split is used only after the model configuration is fixed. No target-day public test data are used for training, hyperparameter search, checkpoint selection, or ensemble selection.

This protocol turns the MC Maze dataset series into a controlled restricted-support cross-day recalibration setting. The comparison with full-data target-day baselines is organized around both the achieved NLB'21 metric and the adaptation cost. Full-data target-day training and interface-only recalibration answer different questions: one estimates the best target-day model under full supervision, while the other tests how much target-day performance can be recovered by reusing a source-day backbone and updating only the neuron interface.

%% file: sections/method.tex
\section{Method}

\begin{figure*}[!t]
    \centering
    \includegraphics[width=\textwidth]{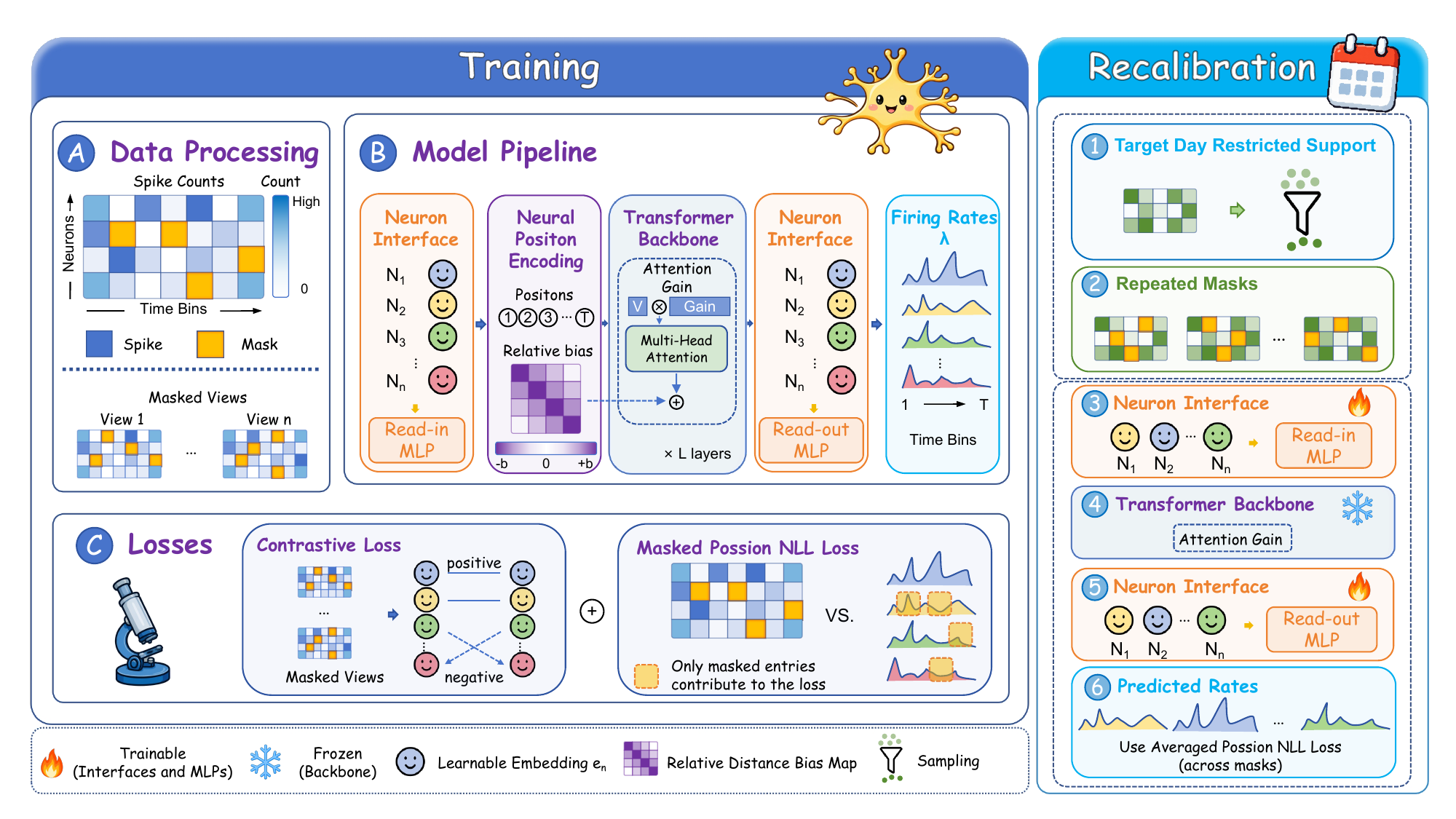}
    \caption{Overview of \method{}. \method{} uses a reusable Transformer dynamics backbone with a neuron interface for read-in/readout and auxiliary attention gain and positional mechanisms for neural activity modeling. Source-day training fits the backbone, neuron interface, and auxiliary mechanisms from masked neural population activity; target-day recalibration freezes the backbone and auxiliary mechanisms and updates only the neuron interface using repeated masking on a small support set.}
    \label{fig:graft-overview}
\end{figure*}

\paragraph{Transformer Dynamics Backbone.} \method{} follows masked neural reconstruction used in Transformer-based neural data models \cite{Ye2021NDT,Le2022STNDT}. Given a masked input $\tilde{\mathbf{Y}}$, the model predicts firing-rate parameters for masked positions. For a mask set $\Omega$, the main loss is the Poisson negative log-likelihood
\begin{equation}
    \mathcal{L}_{\mathrm{mask}}
    =
    -\sum_{(t,n)\in\Omega}
    \log p(y_{t,n}\mid \hat{\lambda}_{t,n}).
    \label{eq:mask-loss}
\end{equation}
The loss is applied to firing-rate parameters rather than to deterministic spike-count reconstruction, matching the NLB'21 evaluation target.

The Transformer backbone receives a sequence of fixed-dimensional representations $\mathbf{x}_{1:T}$ after read-in. It uses self-attention over time bins within a trial to integrate observed neural context and produces contextual states $\mathbf{s}_{1:T}$. The backbone is not directly parameterized by the number of recorded neurons. Changes in neuron count are handled before and after the backbone by the neuron interface.

\paragraph{Neuron Interface.} The neuron interface assigns each recorded neuron a learnable embedding $\mathbf{e}_n$. This embedding represents the neuron's functional role within the model interface, not a claim of stable biological identity across sessions. The read-in pathway maps the embedding to a gain vector
\begin{equation}
    \mathbf{g}_n = \phi_{\mathrm{in}}(\mathbf{e}_n),
    \label{eq:readin-gain}
\end{equation}
where $\phi_{\mathrm{in}}$ is a learnable mapping and $\mathbf{g}_n\in\mathbb{R}^{D}$. For each time bin, spike counts are modulated by these neuron-specific gain vectors and pooled over neurons:
\begin{equation}
    \mathbf{r}_t
    =
    \frac{1}{\sqrt{N}}
    \sum_{n=1}^{N}
    \tilde{y}_{t,n}\mathbf{g}_n .
    \label{eq:readin-pooling}
\end{equation}
The same spike count therefore enters the shared backbone differently depending on which neuron produced it. The factor $1/\sqrt{N}$ stabilizes the scale of the pooled representation when the number of recorded neurons changes.

The readout pathway is also conditioned on the neuron embedding. It generates a readout direction and bias,
\begin{equation}
    \mathbf{o}_n = \phi_{\mathrm{out}}(\mathbf{e}_n),
    \qquad
    b_n = \psi(\mathbf{e}_n),
    \label{eq:readout-params}
\end{equation}
and maps each backbone state $\mathbf{s}_t$ to the log-rate for neuron $n$:
\begin{equation}
    z_{t,n} = \mathbf{s}_t^\top \mathbf{o}_n + b_n,
    \qquad
    \hat{\lambda}_{t,n} = \exp(z_{t,n}).
    \label{eq:readout-rate}
\end{equation}
Thus, the read-in and readout maps are organized around neuron embeddings rather than around a fixed input coordinate system. In target-day recalibration, new target-day neuron embeddings and read-in/readout mappings can be optimized while the temporal backbone remains unchanged.

\paragraph{Temporal and Contextual Modulation.} Attention gain modulates how contextual information enters the Transformer backbone. For each modulated attention layer, a trial-level context vector $\mathbf{c}$ is computed from an early hidden sequence by temporal averaging. A small gain branch maps $\mathbf{c}$ to head-specific value gains,
\begin{equation}
    \boldsymbol{\gamma}^{V,(l)}_m
    =
    \mathbf{1}
    +
    \beta
    \tanh\left([\phi_V^{(l)}(\mathbf{c})]_m\right),
    \label{eq:attention-gain}
\end{equation}
where $l$ indexes the layer and $m$ indexes the attention head. The value matrix is then rescaled as
\begin{equation}
    \tilde{\mathbf{V}}^{(l)}_m
    =
    \mathbf{V}^{(l)}_m
    \odot
    \boldsymbol{\gamma}^{V,(l)}_m .
    \label{eq:value-gain}
\end{equation}
This operation changes the effective strength of value-side content while leaving query-key matching intact. We use it as a lightweight computational gate, related to gated attention mechanisms \cite{Qiu2025GatedAttention}, rather than as a biological claim.

Neural positional encoding supplies two forms of temporal information. A fixed sinusoidal position vector $\mathbf{p}_t$ is added to the read-in representation,
\begin{equation}
    \mathbf{x}_t = \mathbf{r}_t + \mathbf{p}_t ,
    \label{eq:position-input}
\end{equation}
so the backbone can distinguish trial phases such as preparation and movement. In addition, a relative distance bias is added to attention scores:
\begin{equation}
    b^{(l,m)}_{t,s} = -\alpha_m |t-s|,
    \label{eq:relative-bias}
\end{equation}
following the idea of linear attention bias \cite{Press2022ALiBi}. The absolute term encodes within-trial position, while the distance bias favors local temporal continuity without introducing a new learned position table.

\paragraph{Source-Day Training.} Source-day training optimizes the Transformer backbone, attention gain branch, neuron embeddings, read-in mapping, and readout mapping on MC Maze. The primary objective is Eq.~\eqref{eq:mask-loss}. We also add an interface-level contrastive consistency term. Two independently masked views of the same trial are passed through the shared neuron interface, producing two read-in representations. The contrastive loss treats the two views of the same trial as a positive pair and other trials in the batch as negatives, using a SimCLR-style objective \cite{Chen2020SimCLR}. The source-day loss is
\begin{equation}
    \mathcal{L}
    =
    \mathcal{L}_{\mathrm{mask}}
    +
    \lambda_{\mathrm{contrast}}
    \mathcal{L}_{\mathrm{contrast}} .
    \label{eq:source-loss}
\end{equation}
Because this consistency term is applied before the Transformer backbone, it specifically regularizes the neuron interface under masking perturbations. Checkpoints and ensemble members are selected using validation performance, not public test data.

\paragraph{Neuron-Interface Cross-Day Recalibration.} Target-day recalibration starts from a source-day \method{} model. The Transformer dynamics backbone and attention gain branch are frozen. The target-day neuron embeddings, read-in mapping, and readout mapping are updated on the target-day support set. This update range accounts for $9.21\%$ of the model parameters in our implementation.

Restricted-support target-day training uses repeated masking. For each support batch, we sample $R_{\mathrm{mask}}$ independent mask patterns and average the masked Poisson loss:
\begin{equation}
    \mathcal{L}_{\mathrm{adapt}}
    =
    \frac{1}{R_{\mathrm{mask}}}
    \sum_{r=1}^{R_{\mathrm{mask}}}
    \mathcal{L}_{\mathrm{mask}}(\Omega^{(r)}).
    \label{eq:repeated-mask}
\end{equation}
Repeated masking does not add target-day trials or labels. It increases the coverage of reconstruction targets obtained from the same support trials, which is useful when calibration data are limited. After recalibration, the model is evaluated on the target-day public test split using the same NLB'21 metrics as the source-day model.

%% file: sections/experiments.tex
\section{Experiments}

\paragraph{Dataset and Metrics.} Experiments use the \dataset{} dataset series \cite{Pei2021NLB}. MC Maze is a delayed center-out reaching task with obstacles, recorded from motor-related cortical areas of monkey Jenkins. The task has repeated behavioral conditions and structured preparation and movement periods, making it suitable for evaluating single-trial firing-rate modeling and temporal neural dynamics.

The standard MC Maze dataset is used for source-day modeling. It contains 1721 training trials, 574 validation trials, and 574 public test trials, with 137 visible and 45 held-out neurons. For cross-day recalibration, the scaled MC Maze datasets (Small, Medium, and Large) are treated as target days from the same task family. Their training/validation/test trial counts are $75/25/100$, $188/62/100$, and $375/125/100$, respectively. Their visible/held-out neuron counts are $107/35$, $114/38$, and $122/40$. All four datasets use 140 observed time bins and 40 future time bins.

We report the four NLB'21 metrics used for MC Maze. The primary metric is \cobps{}, which evaluates held-out-neuron firing-rate recovery. Secondary metrics are \fpbps{} for future-bin firing-rate prediction, \psthrtwo{} for condition-averaged firing structure, and \velrtwo{} for behaviorally decodable information in the submitted firing rates. Because these metrics evaluate different targets, we interpret \cobps{} as the main neural population modeling metric and use the others to characterize secondary behavior.

\paragraph{Baselines.} For source-day MC Maze, we compare with public NLB'21 results and reported results from latent-variable, Transformer, state-space, and recent neural modeling methods. These include NDT \cite{Ye2021NDT}, AutoLFADS \cite{Keshtkaran2022AutoLFADS}, STNDT \cite{Le2022STNDT}, S5 \cite{Smith2023S5}, DLFM \cite{Wang2026DLFM}, MINT \cite{Perkins2025MINT}, BAND \cite{Kudryashova2025BAND}, SLDS \cite{Ghahramani2000SwitchingSSM}, GPFA \cite{Yu2009GPFA}, and the NLB'21 spike-smoothing baseline. These methods provide the reference range for within-day MC Maze modeling.

For target-day results, public baselines are trained on the corresponding target-day dataset with full target-day training data. We compare with NDT-U \cite{Mifsud2023NDTU}, MINT, AutoLFADS, NDT, SLDS, GPFA, and spike smoothing. NDT-U is a tuned NDT variant reported for the scaled MC Maze datasets. These baselines provide full-data target-day reference scores; \method{} starts from MC Maze and evaluates whether interface-only recalibration can recover target-day firing-rate estimates with a restricted support set.

\paragraph{Evaluation Protocol.} Source-day models are trained on the MC Maze training split. Validation data are used for checkpoint selection and ensemble member selection. For the ensemble result, individual models produce firing-rate parameters, and the ensemble prediction averages these parameters before evaluation. The public test split is used only after the training configuration, checkpoints, and ensemble selection rule are fixed.

Target-day recalibration uses the repeated-masking loss in Eq.~\eqref{eq:repeated-mask}. For the scaled MC Maze datasets Large, Medium, and Small, the support sets contain $256$, $96$, and $48$ training trials. These correspond to $68.3\%$, $51.1\%$, and $64.0\%$ of the target-day training splits. During recalibration, the Transformer dynamics backbone and attention gain branch are frozen; only the target-day neuron interface, including neuron-specific embeddings and read-in/readout mappings, is updated. This update range is $9.21\%$ of model parameters. Target-day validation data select checkpoints and ensemble members. Target-day public test data are reserved for final evaluation and are not used for gradient updates, hyperparameter search, checkpoint selection, or ensemble selection.

Component ablations are run in two settings. Source-day ablations retrain MC Maze models with one mechanism removed or replaced, then evaluate the selected checkpoint under the same source-day metric pipeline. Cross-day ablations start from the source-day checkpoint corresponding to the target-day single-model result and recalibrate on the scaled Large dataset under matched support and validation conditions. We use the scaled Large dataset for cross-day ablations because it provides enough support trials to evaluate recalibration choices while still requiring source-to-target adaptation.

%% file: sections/results.tex
\section{Results}

\paragraph{MC Maze Modeling under NLB'21.} Table~\ref{tab:source-results} reports MC Maze public test results under the standard NLB'21 protocol. The \method{} single model reaches $0.3742$ \cobps{}, above NDT, AutoLFADS, and the reported STNDT single model. The ensemble reaches $0.3866$ \cobps{}, yielding state-of-the-art performance on the primary \cobps{} metric among public and reported NLB'21 results. This MC Maze result shows that the neuron interface and auxiliary Transformer mechanisms contribute to neural population activity modeling before any target-day recalibration is applied.

\begin{table}[!ht]
    \centering
    \footnotesize
    \setlength{\tabcolsep}{2.0pt}
    \renewcommand{\arraystretch}{1.08}
    \begin{tabular}{@{}lcccc@{}}
        \toprule
        Method & \cobps{} & \velrtwo{} & \psthrtwo{} & \fpbps{} \\
        \midrule
        STNDT (single) & 0.3691 & 0.8985 & 0.6567 & 0.2505 \\
        STNDT (ensemble) & \underline{0.3862} & 0.9095 & \underline{0.6693} & \textbf{0.2686} \\
        S5 & 0.3823 & 0.9043 & 0.6431 & 0.2581 \\
        DLFM & 0.3779 & 0.9019 & 0.6525 & \textbf{0.2686} \\
        AutoLFADS & 0.3364 & 0.9097 & 0.6360 & 0.2349 \\
        MINT & 0.3304 & 0.9121 & \textbf{0.7496} & 0.2076 \\
        NDT & 0.3229 & 0.8862 & 0.5308 & 0.2206 \\
        BAND & 0.3215 & \textbf{0.9362} & 0.6187 & 0.2241 \\
        GPFA & 0.1872 & 0.6399 & 0.5150 & -- \\
        \midrule
        \method{} (single) & 0.3742 & 0.9030 & 0.6368 & 0.2491 \\
        \method{} (ensemble) & \textbf{0.3866} & \underline{0.9123} & 0.6472 & \underline{0.2655} \\
        \bottomrule
    \end{tabular}
    \caption{MC Maze public test results under the standard NLB'21 protocol. Bold marks the best result for each metric; underline marks the second-best result.}
    \label{tab:source-results}
\end{table}

The secondary metrics clarify the scope of this state-of-the-art claim. Although \method{} is trained for masked firing-rate modeling rather than explicit behavior decoding, its submitted rates remain behavior-informative. The ensemble reaches $0.9123$ \velrtwo{}, slightly above MINT ($0.9121$) and second only to BAND ($0.9362$). This comparison is meaningful because MINT and BAND are designed to emphasize behavior-informative or behavior-aligned neural structure \cite{Perkins2025MINT,Kudryashova2025BAND}. At the same time, BAND has the best \velrtwo{} and MINT has the best \psthrtwo{}. The MC Maze evidence therefore supports state-of-the-art neural population activity modeling on the primary \cobps{} metric, not superiority on every behavioral or condition-averaged metric.

\paragraph{Restricted-Support Cross-Day Recalibration.} Table~\ref{tab:crossday-results} summarizes the target-day recalibration results on the scaled MC Maze datasets. Across target days, \method{} reuses the Transformer backbone trained on MC Maze and updates only $9.21\%$ of model parameters. It exceeds AutoLFADS full-data \cobps{} on all three target days while using restricted support sets. On Large, \method{} reaches $0.3749$ \cobps{} with $256/375$ training trials and obtains the best \fpbps{} among the listed public references. On Medium, it reaches $0.3112$ \cobps{} with $96/188$ training trials and obtains the best \velrtwo{}. On Small, it reaches $0.3152$ \cobps{} with $48/75$ training trials, remaining above AutoLFADS, NDT, SLDS, GPFA, and spike smoothing.

\begin{table*}[!t]
    \centering
    \footnotesize
    \setlength{\tabcolsep}{2.0pt}
    \renewcommand{\arraystretch}{1.08}
    \begin{tabular}{@{}lcccccccccccc@{}}
        \toprule
        \multirow{2}{*}{Method} & \multicolumn{4}{c}{Large} & \multicolumn{4}{c}{Medium} & \multicolumn{4}{c}{Small} \\
        \cmidrule(lr){2-5}\cmidrule(lr){6-9}\cmidrule(l){10-13}
        & \cobps{} & \velrtwo{} & \psthrtwo{} & \fpbps{} & \cobps{} & \velrtwo{} & \psthrtwo{} & \fpbps{} & \cobps{} & \velrtwo{} & \psthrtwo{} & \fpbps{} \\
        \midrule
        NDT-U & \textbf{0.3849} & 0.9157 & 0.7141 & 0.2004 & \textbf{0.3358} & \underline{0.9026} & \underline{0.6539} & \textbf{0.1819} & \textbf{0.3711} & 0.8490 & \underline{0.5339} & 0.0929 \\
        MINT & \underline{0.3847} & \textbf{0.9330} & \textbf{0.8406} & 0.1961 & \underline{0.3331} & 0.8958 & \textbf{0.7285} & \underline{0.1807} & \underline{0.3463} & \textbf{0.8711} & \textbf{0.5872} & \textbf{0.1563} \\
        AutoLFADS & 0.3740 & 0.9178 & 0.7261 & \underline{0.2039} & 0.3036 & 0.8680 & 0.5969 & 0.1636 & 0.2854 & 0.7982 & 0.3342 & 0.1244 \\
        NDT & 0.3592 & 0.8834 & 0.6248 & 0.1867 & 0.2412 & 0.7206 & 0.2746 & 0.1061 & 0.2368 & 0.6421 & 0.2579 & 0.0765 \\
        SLDS & 0.3031 & 0.7954 & 0.6309 & 0.0370 & 0.2232 & 0.7557 & 0.5163 & 0.0203 & 0.2491 & 0.6324 & 0.4255 & -0.1657 \\
        GPFA & 0.2385 & 0.5765 & 0.5068 & -- & 0.1725 & 0.5664 & 0.4112 & -- & 0.2140 & 0.5138 & 0.4173 & -- \\
        Spike smoothing & 0.2245 & 0.5772 & -0.1189 & -- & 0.1673 & 0.5220 & -0.0676 & -- & 0.1914 & 0.4646 & 0.0401 & -- \\
        \midrule
        \method{} (single) & 0.3671 & 0.9205 & 0.6567 & 0.1953 & 0.3028 & 0.8906 & 0.4030 & 0.1319 & 0.2329 & 0.7829 & 0.2126 & 0.0385 \\
        \method{} (ensemble) & 0.3749 & \underline{0.9309} & \underline{0.7377} & \textbf{0.2120} & 0.3112 & \textbf{0.9090} & 0.5153 & 0.1543 & 0.3152 & \underline{0.8667} & 0.4287 & \underline{0.1267} \\
        \bottomrule
    \end{tabular}
    \caption{Restricted-support cross-day recalibration from MC Maze to scaled MC Maze datasets. Large, Medium, and Small correspond to 2009-10-06, 2009-09-29, and 2009-09-28, respectively. Public baselines use full target-day training data. \method{} uses $256/375$, $96/188$, and $48/75$ target-day support trials on Large, Medium, and Small, respectively, and updates only the neuron interface. Bold marks the best result within each target day and metric; underline marks the second-best result.}
    \label{tab:crossday-results}
\end{table*}

These target-day results characterize data- and parameter-efficient recalibration rather than full-data target-day training. NDT-U and MINT remain stronger than \method{} on \cobps{} for the scaled target days, especially on Small. This gap is informative: with only $48$ support trials, the target-day interface has limited coverage of task conditions and neural population relationships. The result shows that a source-day backbone can support useful target-day firing-rate estimation when the interface is recalibrated under matched task conditions.

\paragraph{Component Ablations.} Table~\ref{tab:ablations} tests whether the proposed mechanisms contribute beyond adding parameters. MC Maze ablations show that replacing the neuron interface with a linear read-in, count embedding, or raw count input lowers \cobps{}. Removing attention gain, neural positional encoding, or interface-level contrastive consistency also lowers \cobps{}. These drops are modest in absolute size, but they are consistent across independent mechanisms and occur under the primary \cobps{} metric.

\begin{table}[!ht]
    \centering
    \footnotesize
    \setlength{\tabcolsep}{0.8pt}
    \renewcommand{\arraystretch}{1.08}
    \begin{tabular}{@{}p{0.43\columnwidth}cccc@{}}
        \toprule
        Setting & $\Delta$\cobps{} & \velrtwo{} & \psthrtwo{} & \fpbps{} \\
        \midrule
        \multicolumn{5}{l}{\textit{A. Embedder mechanism comparison}} \\
        Full \method{} & -- & 0.9030 & 0.6369 & 0.2491 \\
        Linear read-in & -0.0048 & 0.8997 & 0.5829 & 0.2440 \\
        Count embedding & -0.0059 & 0.9000 & 0.6098 & 0.2441 \\
        Raw count input & -0.0077 & 0.9070 & 0.6232 & 0.2518 \\
        \midrule
        \multicolumn{5}{l}{\textit{B. Auxiliary mechanism removal}} \\
        Attention gain & -0.0050 & 0.9061 & 0.6431 & 0.2516 \\
        Neural positional encoding & -0.0050 & 0.9005 & 0.6032 & 0.2527 \\
        Interface contrast & -0.0069 & 0.9016 & 0.6152 & 0.2482 \\
        \midrule
        \multicolumn{5}{l}{\textit{C. Cross-day recalibration ablation}} \\
        $R_{\mathrm{mask}}=4$ & -- & 0.9205 & 0.6567 & 0.1953 \\
        $R_{\mathrm{mask}}=1$ & -0.0178 & 0.8986 & 0.6473 & 0.1764 \\
        $R_{\mathrm{mask}}=2$ & -0.0196 & 0.9017 & 0.6520 & 0.1807 \\
        Freeze read-in/out MLP & -0.0446 & 0.8841 & 0.6490 & 0.1785 \\
        \bottomrule
    \end{tabular}
    \caption{Component ablations. A compares MC Maze embedder mechanisms, B removes auxiliary MC Maze mechanisms, and C ablates MC Maze-to-Large recalibration. For A and B, $\Delta$ is measured relative to Full \method{}; for C, it is measured relative to $R_{\mathrm{mask}}=4$.}
    \label{tab:ablations}
\end{table}

The cross-day ablations make the role of the interface more direct. Reducing repeated masking from $R_{\mathrm{mask}}=4$ to $1$ or $2$ lowers \cobps{} by about $0.018$--$0.020$, suggesting that multiple mask views provide useful calibration signal from the same support trials. Freezing read-in and readout mappings leaves only per-neuron interface embeddings to adapt and reduces \cobps{} by $0.0446$. Cross-day drift is therefore not absorbed solely by assigning new neuron identities; the pathways by which target-day neurons enter and leave the shared backbone also need recalibration.

%% file: sections/discussion.tex
\section{Discussion and Limitations}

\paragraph{Modeling Evidence.} The results support interface-backbone separation as a mechanism for both Transformer-based neural population activity modeling and cross-day recalibration. On MC Maze under the standard NLB'21 protocol, \method{} reaches state-of-the-art \cobps{} among public and reported results, showing that the neuron interface and auxiliary mechanisms contribute to within-day firing-rate estimation. The MC Maze ablations further suggest that neuron-conditioned read-in, attention gain, neural positional encoding, and interface-level contrastive consistency each contribute to the primary \cobps{} metric.

The cross-day experiments add a complementary adaptation result. When the task family is comparable and a source-day backbone has been trained on MC Maze, target-day firing-rate estimation can be recovered to a useful level on scaled MC Maze datasets by recalibrating only the neuron interface. Repeated masking and read-in/readout recalibration are both important under this protocol. This result is consistent with the view that some temporal structure can be reused across recording days, while neuron-specific entry and exit pathways must be adjusted.

\paragraph{Scope of the Evidence.} The learned backbone states should be interpreted through the NLB'21 firing-rate modeling objective rather than as direct biological neural dynamics. Stronger neuroscientific claims would require analyses of latent trajectory geometry, neuron-interface structure, or perturbation responses beyond the benchmark metrics used here.

The cross-day protocol uses MC Maze datasets from the same monkey and task family, with target days close to the source day. Larger changes in behavior, brain area, recording hardware, or time span may require additional adaptation mechanisms. The Small target-day result shows this boundary: with very few support trials, \method{} remains above several full-data baselines but is still below the best tuned target-day methods on \cobps{}.

The BCI motivation also points beyond the present offline evaluation. A deployment-oriented evaluation would test interface-only recalibration under online constraints, including latency, closed-loop feedback, and continual recording changes, where firing-rate accuracy must translate into stable behavior-facing control.

\paragraph{Broader Evaluation.} A broader evaluation would include more NLB'21-style datasets and longer multi-day recordings where task structure, brain area, and recording drift vary more substantially. The learned neuron-interface parameters and readout directions also provide a concrete object for analysis: if the interface captures interpretable neuron groups or recording-day shifts, it could connect the modeling gains reported here to more mechanistic neural analyses. A behavior-facing extension would combine the current firing-rate modeling objective with decoding-oriented objectives to test whether the same interface-backbone separation improves robust BCI control.

%% file: sections/conclusion.tex
\section{Conclusion}

\method{} separates reusable Transformer-based temporal dynamics from a recalibratable neuron interface for neural population activity modeling. The neuron interface controls neuron-specific read-in/readout, while auxiliary gain and positional mechanisms support activity modeling inside the shared Transformer. On MC Maze under the standard NLB'21 protocol, \method{} sets a new state of the art on \cobps{} among public and reported results; on scaled MC Maze datasets, it supports cross-day recalibration by updating only the neuron interface. These results show that interface-backbone separation can support both strong Transformer-based neural population activity modeling and data-efficient cross-day recalibration. For basic neuroscience, this separation provides a way to model population-level temporal structure while making neuron-specific recording changes explicit at the interface. For BCI engineering, it points to recalibration procedures that preserve a source-day temporal model and update only the neuron-specific entry and exit mappings needed for a new recording day.

%% file: sections/appendix.tex
\section{Appendix}

\paragraph{Evaluation Transparency and Test-Set Overfitting Risk.} NLB'21 originally evaluated public test performance through EvalAI, which kept the test targets private and computed metrics from online submissions \cite{Pei2021NLB}. The NLB'21 organizers later stopped accepting EvalAI submissions after January 2026 and released local evaluation data, while noting that public test access increases the risk of test-set overfitting \cite{NLB2026ChallengeInfo}. This change does not invalidate public test results, but it makes the evaluation boundary important for interpreting state-of-the-art claims.

Our training and model selection do not use the public test split. Source-day parameter updates use the MC Maze training split. Validation data select hyperparameters, checkpoints, and ensemble members. Target-day recalibration uses only the target-day support trials for gradient updates, with target-day validation data used for checkpoint and ensemble selection. Public test data are used only after these choices have been fixed. In particular, the source-day ensemble size is not selected by sweeping $K$ on the public test split; we use the validation-ranked candidate pool and report a stable ensemble setting.

To make this boundary auditable, Table~\ref{tab:appendix-overfit} reports two sensitivity checks from the source-day candidate pool. First, selecting ensemble members by validation \cobps{} tolerance gives almost identical \cobps{} under $0.5\%$ and $1.0\%$ relative tolerance. Second, fixed ensemble sizes across a broad range keep public-test \cobps{} near $0.3866$. These checks do not remove the general risk introduced by public test access, but they show that the reported MC Maze result is not the outcome of a narrow test-set-tuned ensemble size.

\begin{table}[htb]
    \centering
    \small
    \setlength{\tabcolsep}{5pt}
    \caption{Source-day MC Maze ensemble sensitivity checks for test-set overfitting risk. All selection rules are defined from validation information before public-test evaluation.}
    \label{tab:appendix-overfit}
    \begin{tabular}{lcccc}
        \hline
        Selection rule & \cobps{} $\uparrow$ & \velrtwo{} $\uparrow$ & \psthrtwo{} $\uparrow$ & \fpbps{} $\uparrow$ \\
        \hline
        Validation \cobps{} tolerance $0.5\%$ & 0.3865 & 0.9122 & 0.6431 & 0.2653 \\
        Validation \cobps{} tolerance $1.0\%$ & 0.3865 & 0.9126 & 0.6489 & 0.2659 \\
        Fixed ensemble size $K=64$ & 0.3865 & 0.9120 & 0.6441 & 0.2654 \\
        Fixed ensemble size $K=76$ & 0.3866 & 0.9121 & 0.6456 & 0.2655 \\
        Fixed ensemble size $K=96$ & 0.3866 & 0.9121 & 0.6463 & 0.2656 \\
        Fixed ensemble size $K=112$ & 0.3866 & 0.9122 & 0.6473 & 0.2656 \\
        Fixed ensemble size $K=128$ & 0.3866 & 0.9123 & 0.6472 & 0.2655 \\
        \hline
    \end{tabular}
\end{table}

This analysis supports a transparency claim rather than recreating the original private EvalAI setting. Within the candidate pool produced by validation-based training and selection, the MC Maze state-of-the-art \cobps{} result is insensitive to reasonable validation-based ensemble rules and to a wide range of ensemble sizes.